\documentclass[conference, 9.5pt]{IEEEtran}
\IEEEoverridecommandlockouts
\usepackage{cite}
\usepackage{amsmath,amssymb,amsfonts}
\usepackage{algorithmic}
\usepackage{graphicx}
\usepackage{textcomp}
\usepackage{xcolor}
\usepackage{flushend}
\usepackage{multirow}
\usepackage[utf8]{inputenc}
\usepackage[vietnam,english]{babel}
\def\BibTeX{{\rm B\kern-.05em{\sc i\kern-.025em b}\kern-.08em
    T\kern-.1667em\lower.7ex\hbox{E}\kern-.125emX}}
\begin{document}

\title{State-of-the-Art Vietnamese Word Segmentation}

\author{\IEEEauthorblockN{Song Nguyen Duc Cong}
\IEEEauthorblockA{Computer Science Department \\
Assumption University\\
Bangkok, Thailand \\
st5429822@au.edu}
\and
\IEEEauthorblockN{Quoc Hung Ngo}
\IEEEauthorblockA{University of Information Technology \\
Ho Chi Minh City, Vietnam\\
hungnq@uit.edu.vn}
\and
\IEEEauthorblockN{Rachsuda Jiamthapthaksin}
\IEEEauthorblockA{Computer Science Department \\
Assumption University\\
Bangkok, Thailand \\
rachsuda@scitech.au.edu}
}

\maketitle

\begin{abstract}
Word segmentation is the first step of any tasks
in Vietnamese language processing. This paper reviews state-of-the-art approaches and systems for word segmentation in
Vietnamese. To have an overview of all stages from building
corpora to developing toolkits, we discuss building the corpus
stage, approaches applied to solve the word segmentation and
existing toolkits to segment words in Vietnamese sentences. In
addition, this study shows clearly the motivations on building
corpus and implementing machine learning techniques to improve
the accuracy for Vietnamese word segmentation. According to
our observation, this study also reports a few of achivements and
limitations in existing Vietnamese word segmentation systems.
\end{abstract}

\begin{IEEEkeywords}
Vietnamese word segmentation; text modelling; Vietnamese corpus
\end{IEEEkeywords}

\section{Introduction}

Lexical analysis, syntactic analysis, semantic analysis, disclosure analysis and pragmatic analysis are five main steps in
natural language processing \cite{b1}, \cite{b2}. While morphology is a
basic task in lexical analysis of English, word segmentation
is considered a basic task in lexical analysis of Vietnamese
and other East Asian languages processing. This task is to
determine borders between words in a sentence. In other
words, it is segmenting a list of tokens into a list of words
such that words are meaningful.

Word segmentation is the primary step in prior to other
natural language processing tasks i. e., term extraction and
linguistic analysis (as shown in Figure 1). It identifies the
basic meaningful units in input texts which will be processed
in the next steps of several applications. For named entity recognization \cite{b3}, word segmentation chunks sentences in input
documents into sequences of words before they are further
classified in to named entity classes. For Vietnamese language,
words and candidate terms can be extracted from Vietnamese
copora (such as books, novels, news, and so on) by using a
word segmentation tool. Conformed features and context of
these words and terms are used to identify named entity tags,
topic of documents, or function words. For linguistic analysis,
several linguistic features from dictionaries can be used either
to annotating POS tags or to identifying the answer sentences.
Moreover, language models can be trained by using machine
learning approaches and be used in tagging systems, like the
named entity recognization system of Tran et al. \cite{b3}.

Many studies forcus on word segmentation for Asian
languages, such as: Chinese, Japanese, Burmese (Myanmar)
and Thai \cite{b4}, \cite{b5}, \cite{b6}, \cite{b7}. Approaches for word segmentation
task are variety, from lexicon-based to machine learning-based
methods. Recently, machine learning-based methods are used
widely to solve this issue, such as: Support Vector Machine or
Conditional Random Fields \cite{b8}, \cite{b9}. In general, Chinese is a
language which has the most studies on the word segmentation
issue. However, there is a lack of survey of word segmentation
studies on Asian languages and Vietnamese as well. This paper
aims reviewing state-of-the-art word segmentation approaches
and systems applying for Vietnamese. This study will be a
foundation for studies on Vietnamese word segmentation and
other following Vietnamese tasks as well, such as part-of-speech tagger, chunker, or parser systems.

\begin{figure}[ht]
    \centering
    \caption{Word segmentation tasks (blue) in the Vietnamese natural lanuage
processing system.}
    \scalebox{0.45}{
        \includegraphics{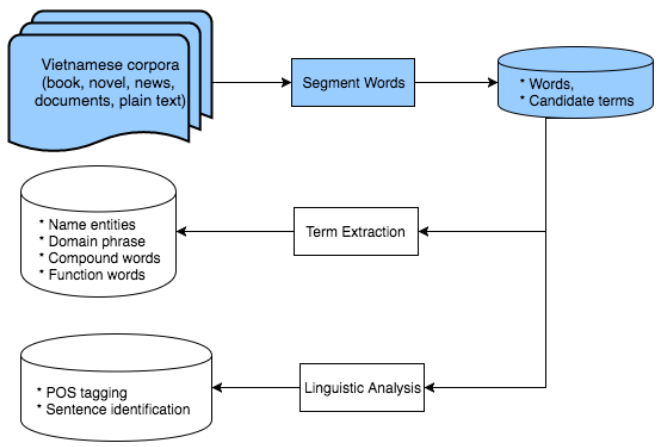}}
    \label{figWSSystem}
\end{figure}

There are several studies about the Vietnamese word segmentation task over the last decade. Dinh et al. started this task
with Weighted Finite State Transducer (WFST) approach and
Neural Network approach \cite{b10}. In addition, machine learning
approaches are studied and widely applied to natural language
processing and word segmentation as well. In fact, several
studies used support vector machines (SVM) and conditional
random fields (CRF) for the word segmentation task \cite{b8}, \cite{b9}.
Based on annotated corpora and token-based features, studies
used machine learning approaches to build word segmentation
systems with accuracy about 94\%-97\%.

According to our observation, we found that is lacks
of complete review approaches, datasets and toolkits which
we recently used in Vietnamese word segmentation. A all sided review of word segmentation will help next studies on
Vietnamese natural language processing tasks have an up-to-date guideline and choose the most suitable solution for the
task. The remaining part of the paper is organized as follows.
Section II discusses building corpus in Vietnamese, containing
linguistic issues and the building progress. Section III briefly
mentions methods to model sentences and text in machine
learning systems. Next, learning models and approaches for
labeling and segmenting sequence data will be presented in
Section IV. Section V mainly addresses two existing toolkits,
vnTokenizer and JVnSegmenter, for Vietnamese word segmentation. Several experiments based on mentioned approaches
and toolkits are described in Section VI. Finally, conclusions
and future works are given in Section VII.

\section{CORPUS}

\subsection{Language Definition}

Vietnamese, like many languages in continental East Asia,
is an isolating language and one branch of Mon-Khmer language group. The most basic linguistic unit in Vietnamese is
morpheme, similar with syllable or token in English and \textviet{“hình
vị”} (phoneme) or \textviet{“tiếng”} (syllable) in Vietnamese. According
to the structured rule of its, Vietnamese can have about 20,000
different syllables (tokens). However, there are about 8,000
syllables used the Vietnamese dictionaries. There are three
methods to identify morphemes in Vietnamese text \cite{b11}.
\begin{itemize}
    \item Morpheme is the smallest meaningful unit of Vietnamese.
    \item Morpheme is the basic unit of Vietnamese.
    \item Morpheme is the smallest meaningful unit and is not used independently in the syntax factor.
\end{itemize}
    
In computational linguistics, morpheme is the basic unit of
languages as Leonard Bloomfield mentioned for English \cite{b12}.
In our research for Vietnamese, we consider the morpheme as
syllable, called \textviet{“tiếng”} in Vietnamese (as Nguyen’s definition
\cite{b13}).

The next concept in linguistics is word which has fully
grammar and meaning function in sentences. For Vietnamese,
word is a single morpheme or a group of morphemes, which
are fixed and have full meaning \cite{b13}. According to Nguyen,
Vietnamese words are able classified into two types, (1) 1-
syllable words with fully meaning and (2) n-syllables words
whereas these group of tokens are fixed. Vietnamese syllable
is not fully meaningful. However, it is also explained in the
meaning and structure characteristics. For example, the token
\textviet{“kỳ”} in \textviet{“quốc kỳ”} whereas \textviet{“quốc”} means national, \textviet{“kỳ”} means
flag. Therefore, \textviet{“quốc kỳ”} means national flag.

Consider dictionary used for evaluating the corpus, extracting features for models, and evaluating the systems, there
are many Vietnamese dictionaries, however we recommend
the Vietnamese dictionary of Hoang Phe, so called Hoang
Phe Dictionary. This dictionary has been built by a group of
linguistical scientists at the Linguistic Institute, Vietnam. It was
firstly published in 1988, reprinted and extended in 2000, 2005
and 2010. The dictionary currently has 45,757 word items with
15,901 Sino-Vietnamese word items (accounting for 34.75\%)
\cite{b14}.

\subsection{Name Entity Issue}

In Vietnamese, not all of meaningful proper names are in
the dictionary. Identifying proper names in input text are also
important issue in word segmentation. This issue is sometimes
included into unknown word issue to be solved. In addition,
named entity recognition has to classify it into several types
such as person, location, organization, time, money, number,
and so on.

Proper name identification can be solved by characteristics.
For example, systems use beginning characters of proper
names which are uppercase characters. Moreover, a list of
proper names is also used to identify names in the text.
In particular, a list of 2000 personal names extracted from
VietnamGiaPha, and a list of 707 names of locations in
Vietnam extracted from vi.wikipedia.org are used in the study
of Nguyen et al. for Vietnamese word segmentation \cite{b8}.

\subsection{Building Corpus}

In general, building corpus is carried out through four
stages: (1) choose target of corpus and source of raw data;
(2) building a guideline based on linguistics knowledge for
annotation; (3) annotating or tagging corpus based on rule
set in the guideline; and (4) reviewing corpus to check the
consistency issue.

Encoding word segmentation corpus using B-I-O tagset
can be applied, where B, I, and O denoted begin of word,
inside of word, and others, respectively. For example, the
sentence \textviet{“Megabit trên giây là đơn vị đo tốc đọ truyền dẫn dữ
liệu ."} (”Megabit per second is a unit to measure the network
traffic.” in English) with the word boundary result \textviet{“Megabit
trên giây là đơn\_vị đo tốc\_độ truyền\_dẫn dữ\_liệu .}" is encoded
as \textviet{“Megabit/B trên/B giây/B là/B đơn/B vị/I đo/B tốc/B độ/I
truyền/B dẫn/I dữ/B liệu/I ./O"} .

Annotation guidelines can be applied to ensure that annotated corpus has less errors because the manual annotation is
applied. Even though there are guidelines for annotating, the
available output corpora are still inconsistent. For example,
for the Vietnamese Treebank corpus of the VLSP\footnote{http://vlsp.vietlp.org:8080/demo/} project,
Nguyen et al. listed out several Vietnamese word segmentation
inconsistencies in the corpus based on POS information and
n-gram sequences \cite{b15} .

\begin{table}[b]
  \centering
  \label{tableAgFarmNews}%
    \setlength{\tabcolsep}{1pt}
  \caption{VIETNAMESE WORD CORPUS}{
    \begin{tabular}{|c|c|c|c|c|}
      \hline
        \textbf{Corpus} & \textbf{Domain} & \textbf{No. of Articles} & \textbf{No. of Sentences} & \textbf{No. of Words}\\
      \hline
CADASA & General book& 5 & 24,240 & 229,357\\
      \hline
vnQTAG & Short novels & 7 & & 74755 \\
      \hline
EVBNews & \multirow{2}{*}{General news} & \multirow{2}{*}{1,000} & \multirow{2}{*}{45,531} & \multirow{2}{*}{832,441} \\
(EVBCorpus) & & & &\\
      \hline
    \end{tabular}}%
\end{table}%

Currently, there are at least three available word segmentation corpus used in Vietnamese word segmentation studies and
systems. Firstly, Dinh et al. built the CADASA corpus from
CADASA’s books \cite{b16}. Secondly, Nguyen et al. built vnQTAG
corpus from general news articles \cite{b8}. More recently, Ngo
et al. introduced the EVBCorpus corpus, which is collected
from four sources, news articles, books, law documents, and
novels. As a part of EVBCorpus, EVBNews, was annotated
common tags in NLP, such as word segmentation, chunker,
and named entity \cite{b17}. All of these corpora are collected from
news articles or book stories, and they are manually annotated
the word boundary tags (as shown in Table I).

\section{TEXT MODELLING AND FEATURES}

To understand natural language and analyze documents
and text, computers need to represent natural languages as
linguistics models. These models can be generated by using
machine learning methods (as show in Figure 2).
    
\begin{figure}[ht]
    \centering
    \caption{Word segmentation data process}
    \scalebox{0.43}{
        \includegraphics{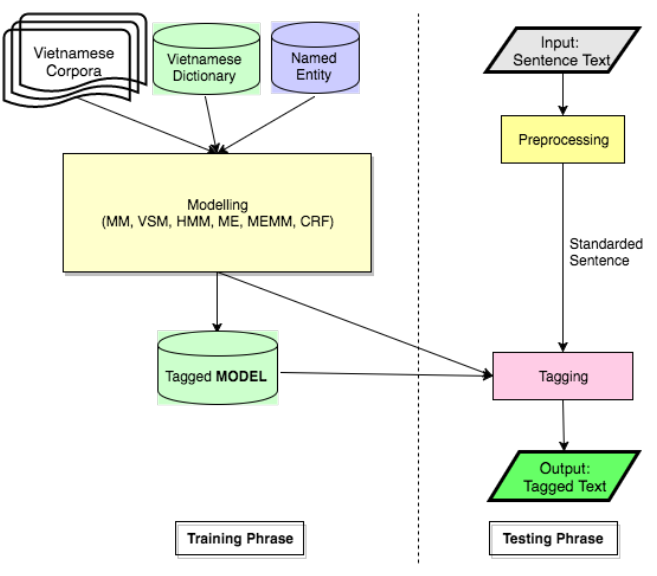}}
    \label{figWSProcess}
\end{figure}
    
There are two common modeling methods for basic NLP
tasks, including n-gram model and bag-of-words model. The
n-gram model is widely used in natural language processing
while the bag-of-words model is a simplified representation
used in natural language processing and information retrieval
\cite{b18}, \cite{b19}. According to the bag-of-words model, the representative vector of sentences in the document does not preserve
the order of the words in the original sentences. It represents
the word using term frequency collected from the document
rather than the order of words or the structure of sentences
in the document. The bag-of-words model is commonly used
in methods of document classification, where the frequency
of occurrence of each word is used as an attribute feature for
training a classifier. In contrast, an n-gram is a contiguous
sequence of \textit{n} items from a given sequence of text. An n-gram model is a type of probabilistic language model for
predicting the next item in a given sequence in form of a
Markov model. To address word segmentation issue, the n-gram model is usually used for approaches because it considers
the order of tokens in the original sentences. The sequence is
also kept the original order as input and output sentences.

\section{BUILDING MODEL METHODS}

There are several studies for Vietnamese Word Segmentation during last decade. For instance, Dinh et al. started the
word segmentation task for Vietnamese with Neural Network
and Weighted Finite State Transducer (WFST) \cite{b10}. Nguyen et
al. continued with machine learning approaches, Conditional
Random Fields and Support Vector Machine \cite{b8}. Most of
statistical approaches are based on the architecture as shown
in Figure 2. According to the architecture, recent studies and
systems focus on either improving or modifying difference
learning models to get the highest accuracy. Features used in
word segmentation systems are syllable, dictionary, and entity
name. The detail of all widely used techniques applied are
collected and described in following subsections.

\subsection{Maximum Matching}

Maximum matching (MM) is one of the most popular
fundamental and structural segmentation algorithms for word
segmentation \cite{b20}. This method is also considered as the
Longest Matching (LM) in several research \cite{b10}, \cite{b4}. It is
used for identifying word boundary in languages like Chinese,
Vietnamese and Thai. This method is a greedy algorithm,
which simply chooses longest words based on the dictionary.
Segmentation may start from either end of the line without any
difference in segmentation results. If the dictionary is sufficient
\cite{b20}, the expected segmentation accuracy is over 90\%, so it is
a major advantage of maximum matching . However, it does
not solve the problem of ambiguous words and unknown words
that do not exist in the dictionary.

There are two types of the maximum matching approach:
\textbf{forward MM (FMM)} and \textbf{backward MM (BMM)}. FMM
starts from the beginning token of the sentence while BMM
starts from the end. If the sentence has word boundary ambiguities, the output of FMM and BMM will be different. When
applying FMM and BMM, there are two types of common
errors due to two ambiguities: \textbf{overlapping ambiguities} and
\textbf{combination ambiguity}. Overlapping ambiguities occur when
the text AB has both word A, B and AB, which are in the
dictionary while the text ABC has word AB and BC, which are
in the dictionary. For example, \textviet{"cụ già đi nhanh quá"} (there two
meanings: ”the old man goes very fast” or ”the old man died
suddenly”) is a case of the overlapping ambiguity while \textviet{"tốc
độ truyền thông tin"} is a case of the combination ambiguity.

As shown in Figure 2, the method simplification ambiguities, maximum matching is the first step to get features for the
modelling stage in machine learning systems, like Conditional
Random Fields or Support Vector Machines.

\subsection{Hidden Markov Model (HMM)}

In Markov chain model is represented as a chain of tokens
which are observations, and word taggers are represented as
predicted labels. Many researchers applied Hidden Markov
model to solve Vietnamese word segmentation such as in \cite{b9},
\cite{b21} and so on.

N-gram language modeling applied to estimate probabilities for each word segmentation solution \cite{b22}. The result of
this method depends on copora and is based maximal matching
strategy. So, they do not solve missing word issue. Let \(P(s)\)
is a product of probabilities of words created from sentence s
(1) with length \(m\) :
\begin{equation}
    s=w_1...w_2
\end{equation}
Each conditional probability of word is based on the last \textit{n-1} words (n-gram) in the sentence s. It is estimated by Markov chain model for word \textit{w} from position \textit{i-n+1} to i-1 with
probability (2)
\begin{equation}
    P(w_i\mid w_{i-n+1}^{i-1})
\end{equation}
We have equation (3)
\begin{equation}
    P(s) = \prod_{i=1}^{m}
    P(w_i\mid w_{i}^{i-1})
    \approx 
    \prod_{i=1}^{m}
    P(w_i\mid w_{i-n+1}^{i-1})
\end{equation}
\subsection{Maximum Entropy (ME)}

Maximum Entropy theory is applied to solve Vietnamese
word segmentation \cite{b16}, \cite{b23}, \cite{b24}. Some researchers do not
want the limit in Markov chain model. So, they use the context
around of the word needed to be segmented. Let \textit{h} is a context,
\textit{w} is a list of words and \textit{t} is a list of taggers, Le \cite{b16}, \cite{b23}
used
\begin{equation}
    P(t_1...t_n\mid w_1...w_n)
    \approx 
    \prod_{i=1}^{n}
    P(t_i\mid h_i)
\end{equation}
\textit{P(s)} is also a product of probabilities of words created from
sentence \(s\) (1). Each conditional probability of word is based
on context \textit{h} of the last \textit{n} word in the sentence s.

\subsection{Conditional Random Fields}

To tokenize a Vietnamese word, in HMM or ME, authors
only rely on features around a word segment position. Some
other features are also affected by adding more special attributes, such as, in case ’?’ question mark at end of sentence,
Part of Speech (POS), and so on. Conditional Random Fields
is one of methods that uses additional features to improve the
selection strategy \cite{b8}.

There are several CRF libraries, such as CRF++\footnote{https://taku910.github.io/crfpp/}, CRFsuite\footnote{http://www.chokkan.org/software/crfsuite/}. These machine learning toolkits can be used to solve the
task by providing an annotated corpus with extracted features.
The toolkit will be used to train a model based on the corpus
and extract a tagging model. The tagging model will then be
used to tag on input text without annotated corpus. In the
training and tagging stages, extracting features from the corpus
and the input text is necessary for both stages.

\subsection{Support Vector Machines}

Support Vector Machines (SVM) is a supervised machine
learning method which considers dataset as a set of vectors
and tries to classify them into specific classes. Basically, SVM
is a binary classifier. however, most classification tasks are
multi-class classifiers. When applying SVMs, the method has
been extended to classify three or more classes. Particular NLP
tasks, like word segmentation and Part-of-speech task, each
token/word in documents will be used as a feature vector. For
the word segmentation task, each token and its features are
considered as a vector for the whole document, and the SVM
model will classify this vector into one of the three tags (B-IO).

This technique is applied for Vietnamese word segmentation in several studies \cite{b8}, \cite{b25}. Nguyen et al. applied on 
a segmented corpus of 8,000 sentences and got the result
at 94.05\% while Ngo et al. used it with 45,531 segmented
sentences and get the result at 97.2\%. It is worth to mention
that general SVM libraries (such as LIBSVM\footnote{https://www.csie.ntu.edu.tw/ cjlin/libsvm/}, LIBLINEAR\footnote{https://www.csie.ntu.edu.tw/ cjlin/liblinear/},
SVMlight\footnote{http://svmlight.joachims.org/}, Node-SVM\footnote{https://github.com/nicolaspanel/node-svm}, and TreeSVM\footnote{https://github.com/sitfoxfly/tree-svm} ), YamCha\footnote{http://chasen.org/ taku/software/yamcha/} is an
opened source SVM library that serves several NLP tasks: POS
tagging, Named Entity Recognition, base NP chunking, Text
Chunking, Text Classification and event Word Segmentation.

\section{TOOLKITS}

vnTokenizer and JVnSegmenter are two famous segmentation toolkits for Vietnamese word segmentation. Both two
word segmentation toolkits are implemented the word segmentation data process in Figure 2. This section gives more details
of these Vietnamese word toolkits.

\subsection{Programming Languages}

In general, Java and C++ are the most common language
in developing toolkits and systems for natural language processing tasks. For example, GATE\footnote{https://gate.ac.uk/}, OpenNLP\footnote{https://opennlp.apache.org/}, Stanford
CoreNLP\footnote{http://stanfordnlp.github.io/CoreNLP/} and LingPipe\footnote{http://alias-i.com/lingpipe/} platforms are developed by JAVA
while foundation tasks and machine learning toolkits are developed by C++. CRF++\footnote{https://taku910.github.io/crfpp/}, SVMLight\footnote{http://svmlight.joachims.org/} and YAMCHA\footnote{http://chasen.org/ taku/software/yamcha/} . Recently, Python becomes popular among the NLP community.
In fact, many toolkits and platforms have been developed by
this language, such as NLTK\footnote{http://www.nltk.org/}, PyNLPl\footnote{https://github.com/proycon/pynlpl} library for Natural
Language Processing.

\subsection{JVnSegmenter}

JVnSegmenter\footnote{http://jvnsegmenter.sourceforge.net/} is a Java-based Vietnamese Word Segmentation Tool developed by Nguyen and Phan. The segmentation model in this tool was trained on about 8,000
tagged Vietnamese text sentences based on CRF model and
the model extracted over 151,000 words from the training
corpus. In addition, this is used in building the EnglishVietnamese Translation System \cite{b26}, \cite{b27}, \cite{b28}. Vietnamese
text classification \cite{b29} and building Vietnamese corpus \cite{b30},
\cite{b31}.

\subsection{vnTokenizer}

vnTokenizer\footnote{http://mim.hus.vnu.edu.vn/phuonglh/softwares/vnTokenizer} is implemented in Java and bundled as
Eclipse plug-in, and it has already been integrated into vnToolkit, an Eclipse Rich Client application, which is intended
to be a general framework integrating tools for processing of
Vietnamese text. vnTokenizer plug-in, vnToolkit and related
resources, including the lexicon and test corpus are freely
available for download. According to our observation, many
research cited vnTokenizer\footnote{https://scholar.google.com/} to use word segmentation results
for applications as building a large Vietnamese corpus \cite{b32},
building an English-Vietnamese Bilingual Corpus for Machine
Translation \cite{b33}, Vietnamese text classification \cite{b34}, \cite{b35}, etc.

\section{EVALUATION AND RESULTS}

This research gathers the results of Vietnamese word segmentation of several methods into one table as show in Table
II. It is noted that they are not evaluated on a same corpus. The
purpose of the result illustration is to provide an overview of
the results of current Vietnamese word segmentation systems
based on their individual features. All studies mentioned in the
table have accuracy around 94-97\% based on their provided
corpus.

This study also evaluates the Vietnamese word segmentation based on existing toolkits using the same annotated
Vietnamese word segmentation corpus. There are two available
toolkits to evaluate and to segment. To be neutral to both
toolkits, we use the EVBNews Vietnamese corpus, a part of
EVBCorpus, to evaluate Vietnamese word segmentation. The
EVBNews corpus contains over 45,000 segmented Vietnamese
sentences extracted from 1,000 general news articles (as shown
in Table III) \cite{b17}. We used the same training set which has
1000 files and 45,531 sentences. vnTokenizer outputs 831,455
Vietnamese words and 1,206,475 tokens. JVnSegmenter outputs 840,387 words and 1,201,683. We correct tags (BIO), and
compare to previous outputs, we have rate from vnTokenizer
is 95.6\% and from JVnsegmenter is 93.4\%. The result of
both vnTokenizer and JVnSegmenter testing on the EVBNews
Vietnamese Corpus are provided in Table IV.

\section{CONCLUSIONS AND FUTURE WORKS}

This study reviewed state-of-the-art approaches and systems of Vietnamese word segmentation. The review pointed
out common features and methods used in Vietnamese word
segmentation studies. This study also had an evaluation of the
existing Vietnamese word segmentation toolkits based on a
same corpus to show advantages and disadvantages as to shed
some lights on system enhancement.

There are several challenges on supervised learning approaches in future work. The first challenge is to acquire
very large Vietnamese corpus and to use them in building a
classifier, which could further improve accuracy. In addition,
applying linguistics knowledge on word context to extract
useful features also enhances prediction performance. The
second challenge is design and development of big data
warehouse and analytic framework for Vietnamese documents,
which corresponds to the rapid and continuous growth of
gigantic volume of articles and/or documents from Web 2.0
applications, such as, Facebook, Twitter, and so on. It should
be addressed that there are many kinds of Vietnamese documents, for example, Han - Nom documents and old and
modern Vietnamese documents that are essential and still
needs further analysis. According to our study, there is no a
powerful Vietnamese language processing used for processing
Vietnamese big data as well as understanding such language.
The final challenge relates to building a system, which is able
to incrementally learn new corpora and interactively process
feedback. In particular, it is feasible to build an advance NLP
system for Vietnamese based on Hadoop platform to improve
system performance and to address existing limitations.

\begin{table*}[ht]
\centering
  \label{tablePrevStudies}%
  \caption{EVBNEWS VIETNAMESE CORPUS}{
    \begin{tabular}{lllcc}
      \hline\noalign{\smallskip}
      \textbf{Method}  & \textbf{Features} & \textbf{Corpus} & \textbf{Result} & \textbf{Result}\\
      \hline\noalign{\smallskip}
      \textbf{NN with WFST}	    & Dictionary, proper name &	305 newspaper articles, &	98.36\% &   Dinh et al. \cite{b10}\\
                        & Morphological analyzer & 7,800 sentences & \\
      \hline\noalign{\smallskip}
      \textbf{ME} & Syllable, dictionary, proper name, & CADASA & 94.44\% & Dien Dinh and Thuy Vu \cite{b16}\\
        & misc, Vietnamese Syllable & 24,240 sentences & \\
      \hline\noalign{\smallskip}
        \textbf{ME} & Syllable, dictionary, proper name, & 4,800 sentences, & 93.70\% &  Le et al. \cite{b23}\\
        & misc, Vietnamese Syllable & 113, 000 syllables 22 \\
      \hline\noalign{\smallskip}
        \textbf{SVMs} & Syllable, BMM, FMM, proper & 1000 newspaper articles,& 97.2\% & Ngo et al. \cite{b36} \\
        & name, misc, foreign & 45,531 sentences  \\
        & & 768,031 Vietnamese words \\
      \hline\noalign{\smallskip}
        \textbf{SVMs} & Syllable, dictionary, proper name, & 305 newspaper articles, & 94.05\% & Nguyen et al. \cite{b8}\\
        & misc, Vietnamese Syllable & 150 novel sentences  \\
        & & 8,000 sentences 23 \\
      \hline\noalign{\smallskip}
        CRFs & Syllable, dictionary, proper name, & 305 newspaper articles, & 95\% & Nguyen et al. \cite{b8} \\ 
        & misc, Vietnamese Syllable & 150 novel articles & \\
        & & 8,000 sentences & \\

      \hline\noalign{\smallskip}
    \end{tabular}}%
\end{table*}%

\begin{table}[ht]
  \centering
  \label{tableEVBNews}%
  \caption{EVBNEWS VIETNAMESE CORPUS}{
    \begin{tabular}{cr}
      \hline\noalign{\smallskip}
        & \textbf{Statistics}\\
      \hline\noalign{\smallskip}
      Number of Files &	1,000 \\
      Number of Sentences   &	45,531 \\
      Number of Words &	832,441 \\
      Number of Tokens   &	832,441 \\
      \hline\noalign{\smallskip}
    \end{tabular}}%
\end{table}%

\begin{table}[ht]
  \centering
  \label{tableResult}%
  \caption{VIETNAMESE WORD SEGMENTATION RESULT OF VNTOKENIZER AND JVNSEGMENTER}{
    \begin{tabular}{crr}
      \hline\noalign{\smallskip}
      & \textbf{vnTokenizer} & \textbf{JVnSegmenter}\\
      \hline\noalign{\smallskip}
      Number of Files &	1,000 &	1,000 \\
      Number of Sentences &	45,531   &	45,531 \\
      Number of Words &	831,455 &	840,387 \\
      Number of Tokens &	1,206,475  &	1,201,683 \\
      Correct Tags   &	1,153,198   &	1,122,752 \\
      \hline\noalign{\smallskip}
      Rate    &	95.6\% &	93.4\% \\
      \hline\noalign{\smallskip}
    \end{tabular}}%
\end{table}%

\end{document}